\documentclass[10pt, conference, compsocconf]{IEEEtran}
\usepackage{tikz}

\ifCLASSINFOpdf
\else
\fi
\usepackage[cmex10]{amsmath}
\usepackage{amsmath, amsthm, amssymb, amsfonts}
\usepackage{graphicx}

\newcommand\copyrighttext{%
  \footnotesize \copyright 2018 IEEE. Personal use of this material is permitted. Permission from IEEE must be
obtained for all other uses, in any current or future media, including
reprinting/republishing this material for advertising or promotional purposes, creating new
collective works, for resale or redistribution to servers or lists, or reuse of any copyrighted
component of this work in other works.}
\newcommand\copyrightnoticeb{%
\begin{tikzpicture}[remember picture,overlay]
\node[anchor=south,yshift=10pt] at (current page.south) {\fbox{\parbox{\dimexpr\textwidth-\fboxsep-\fboxrule\relax}{\copyrighttext}}};
\end{tikzpicture}%
}

\theoremstyle{definition}
\newtheorem{definition}{Definition}[section]

\usepackage{secantMACROS}

\newcommand{\R}{\mathbb{R}}                      
\DeclareMathOperator*{\argmax}{arg\,max}

\begin{document}
%
\title{Monitoring the shape of weather, soundscapes, and dynamical systems: a new statistic for dimension-driven data analysis on large data sets}


\author{\IEEEauthorblockN{Henry Kvinge, Elin Farnell, Michael Kirby, Chris Peterson}
\IEEEauthorblockA{Department of Mathematics\\
Colorado State University\\
Fort Collins, CO 80523-1874}
}


%


\maketitle

\copyrightnoticeb

\begin{abstract}
Dimensionality-reduction methods are a fundamental tool in the analysis of large data sets. These algorithms work on the assumption that the ``intrinsic dimension'' of the data is generally much smaller than the ambient dimension in which it is collected. Alongside their usual purpose of mapping data into a smaller dimension with minimal information loss, dimensionality-reduction techniques implicitly or explicitly provide information about the dimension of the data set.

In this paper, we propose a new statistic that we call the $\kappa$-profile for analysis of large data sets. The $\kappa$-profile arises from a dimensionality-reduction optimization problem: namely that of finding a projection into $k$-dimensions that optimally preserves the secants between points in the data set. From this optimal projection we extract $\kappa,$ the norm of the shortest projected secant from among the set of all normalized secants. This $\kappa$ can be computed for any $k$; thus the tuple of $\kappa$ values (indexed by dimension) becomes a $\kappa$-profile. Algorithms such as the Secant-Avoidance Projection algorithm and the Hierarchical Secant-Avoidance Projection algorithm, provide a computationally feasible means of estimating the $\kappa$-profile for large data sets, and thus a method of understanding and monitoring their behavior. As we demonstrate in this paper, the $\kappa$-profile serves as a useful statistic in several representative settings: weather data, soundscape data, and dynamical systems data. 


\end{abstract}


%
\IEEEpeerreviewmaketitle

\section{Introduction}

As high-dimensional data becomes more and more plentiful, dimensionality-reduction algorithms become an increasingly important tool for any researcher seeking to extract meaningful information from their data. Indeed, it is not unusual to find that data collected in an $n$-dimensional space is intrinsically only $k$-dimensional, where $k \ll n$. A good dimensionality-reduction algorithm will find a map which takes the data from $\MB{R}^n$ to $\MB{R}^{k'}$ (for some $k'$ close to $k$) while preserving fundamental properties such as the distances between data points. This process is fundamental since data in low-dimensional space is often computationally easier to store and manipulate, and hence a wider array of algorithms are available for data analytics. Furthermore, the process of reduction often coincides with the production of a more appropriate representation of the data, making many data analytics algorithms more successful as a result of a meaningful feature space.

By systematically studying how well a data set $D$ in $\mathbb{R}^n$ can be projected to $\mathbb{R}^m$ for $m < n$, as $m$ varies, we can begin to uncover basic geometric properties of $D$ (such as the intrinsic dimension of an underlying manifold on which $D$ approximately sits). In this paper we will explore this idea in the context of secant-based dimensionality reduction algorithms. These are a family of dimensionality-reduction algorithms that use the secant set $S$ of $D$ to find projections that best preserve distances between points. As a consequence of producing these projections, the algorithms also return a statistic which we call the $\kappa$-profile. We will focus on $\kappa$-profiles obtained from SAP and HSAP dimensionality-reduction algorithms \cite{kvinge2018gpu,kvinge2018too}.

The problem of calculating the dimension of a data set has been addressed from various perspectives. The classical approach to estimating dimension is to use a linear method such as principal component analysis (PCA). However, such methods do not capture the dimension of non-linear data well. Other methods that do not assume linearity have also been proposed, see for example \cite{camastra2003data,wang2015survey,cunningham2015linear,fukunaga1971algorithm,costa2006determining,kirby_hundley1999,broomhead_jones_king87}.

Though it is related to dimension, the $\kappa$-profile often carries more information. In particular, one may observe gradual transitions in $\kappa,$ whereas an estimate of the intrinsic dimension will by nature either remain constant or have a sudden jump discontinuity. The $\kappa$-profile also gives information about how well the data fits into many different reduced dimensions simultaneously. This is useful for studying real-world data that almost never conforms precisely to a manifold.

The $\kappa$-profile can be used not only for analysis of a static data set, but also for settings such as time series analysis and anomaly detection. In particular, the sensitivity of the $\kappa$-profile to changes in the geometry of a data set makes it a prime candidate as an indicator of fundamental changes in the behavior of an underlying system as a function of either time or a set of parameters. A useful transformation that helps to reveal structure underlying a time series is achieved through a time-delay embedding. In fact, by considering our data via a time-delay embedding, the $\kappa$-profile can in some cases also give indication of the dimension of the underlying dynamics from which the data is drawn.

Estimating the $\kappa$-profile from a data set is relatively easy and requires very few assumptions about the data. It is thus a useful statistic, particularly in cases in which domain-specific tools are limited. 

This paper is organized as follows. In Section \ref{sect-background} we review the mathematical concept of dimension of a dataset, secant-based dimensionality reduction algorithms, and formally define the $\kappa$-profile. We also include a brief review of geodesic distance on a Grassmann manifold in order to provide context for comparison between dimensionality-reduction related to the $\kappa$-profile and dimensionality-reduction via principal component analysis. In Section \ref{sect-dim-time-series} we discuss ways to format a collection of time-parametrized data sets prior to calculating the $\kappa$-profile and provide mathematical justification for this framework. In Section \ref{sect-KS-equation} we discuss a synthetic example where the dimension of the data, the solution set of a well-known partial differential equation, is already approximately known. In \ref{sect-weather} and \ref{sect-soundscapes} we calculate the $\kappa$-profile on weather data and ambient noise data respectively.

\section{Background} \label{sect-background}

We begin by reviewing mathematical prerequisites. 

\subsection{Dimension Estimation} \label{subsect-dim}
We follow the background on dimension estimation from \cite{kvinge2018too}. The idea that motivates the consideration of dimension is that a data set, at least locally, has hidden constraints that allow one to consider the data as a noisy sampling of some underlying manifold.
When we say that a data set is $d$-dimensional, we mean that the data can be viewed as being noisily sampled from a $d$-dimensional manifold. Locally, a $d$-dimensional manifold can be
parameterized by $d$ free variables. The dimension gives one a coarse measure of the complexity of the manifold.
The estimation of dimension from data has been addressed by numerous authors
including, e.g.,  \cite{camastra2003data,wang2015survey,cunningham2015linear,costa2006determining,kirby_hundley1999,AnHuKi02,broomhead_jones_king87,fukunaga1971algorithm,BK05}.

There is a useful theoretical result for characterizing
dimension-preserving transformations.  It revolves around the definition
of a  {\it bi-Lipschitz} function.
A function $f(x)$ is said to be {\it bi-Lipschitz} on $X$ if there exists $a,b > 0$ such that for all $x,y \in X$ 
$$
a \| x - y \|_{\ell_2} \le \| f(x) - f(y) \|_{\ell_2} \le  b \| x - y \|_{\ell_2}.
$$
The constant $a$ restricts pairs of points from collapsing on top of each other, i.e. from distinct points being identified as the same point after the application of the function. The constant $b$ restricts pairs of points from blowing apart. In other words, points that are close together before the application of the function should remain close together after the application of the function.  In the context of projection, which is the class of functions considered in this paper for the estimates of dimension, we can restrict to the case $b=1$. A key feature of bi-Lipschitz functions is:
$$
{\rm if}\ f:X \rightarrow Z \ {\textrm{ is bi-Lipschitz, then}} \ 
\dim(X) =  \dim(Z)
$$
where the dimension can be taken
as the topological dimension, or the Hausdorff dimension; 
see~\cite{Fal03} for details.
Thus we see a link between dimension-preservation and projections that
avoid collapsing secants.  Projection-based algorithms
that maximally avoid decreasing the length of secants are, in some sense, optimally dimension-preserving.
An additional  argument for this approach, based on invoking Whitney's easy embedding theorem, is made in \cite{BK00}.

\subsection{The SAP and HSAP algorithms and $\kappa$-profiles}

Following the context in \cite{kvinge2018gpu}, data residing in a high-dimensional ambient space can both be a computational burden and difficult to analyze. Data-reduction algorithms offer a way to reduce these difficulties by mapping the data set into a lower-dimensional ambient space with the goal of retaining as much information as possible. A classical example of such an algorithm is PCA. In \cite{BK05,BK00,BK01}, Broomhead and Kirby developed a new framework for data reduction which focuses on preserving the normalized secant set 
\begin{equation*}
S := \Big\{\frac{x - y}{||x-y||_2} \;| \; x, y \in D\Big\}
\end{equation*}
corresponding to a data set $D \subset \mathbb{R}^n$. This framework is inspired by the constructive proof of Whitney's embedding theorem, a theorem from differential topology which gives an upper bound on the dimension of Euclidean space required to smoothly embed a compact $k$-dimensional manifold \cite[Section 1.8]{GP10}. When successful, the projections produced by these methods not only retain differential structure but also have a well-conditioned inverse (roughly, an inverse for which small changes in the domain produce similarly small changes in the range). This property, not necessarily found for projections obtained from other popular methods such as PCA, means that the projection provides not only a method to compress the data, but also to decompress the data without loss of information.

In practice, producing projections from a data set $X \subset \R^n$ into $\R^m$ for $m < n$ within the Whitney reduction framework involves finding projections $P:\R^n \rightarrow \R^m$ for which the image of $X$ is a minimal distortion of $X$. As motivated by the bi-Lipschitz discussion above, this amounts to finding projections, $P$, for which the smallest value of $||P(s)||_{\ell_2}$ over all $s \in S$ is maximized. 

In \cite{kvinge2018gpu} the authors propose the Secant-Avoidance Projection (SAP) algorithm for producing a projection that best preserves the secant set of a data set. The algorithm proceeds through an iterative procedure. An outline of the algorithm is as follows. Let $X\subset\R^n$ be a data set and define $S$ to be the set of all normalized secants of $X.$ Initialize a projection from $\R^n$ to $\R^m.$ At iteration $i,$ use the current projection to compute the projection of the secant set $S$ and determine the secant vector that is least well-preserved, i.e. has the shortest projection. Define the ($i+1$)-th projection by rotating the $i$-th projection subspace toward the orthogonal complement of the projection of the shortest projected secant. The $(i+1)$-th projection is the projection corresponding to this rotated $m$-dimensional subspace.

As a means of addressing the challenges of big data, the authors of \cite{kvinge2018too} propose a generalization of SAP that relies on clustering the data. In this case, the algorithm is called HSAP: Hierarchical Secant-Avoidance Projection. The algorithm design is similar but relies on linear approximations to the clusters of data along with principal angles as a measure of closeness.  

The central mathematical object of this paper is what we call a $\mathbf{\kappa}$-profile.
\begin{definition}
Let $X$ be a finite collection of points in $\R^n,$ and let $S$ be the set of normalized secants for $X.$ For a fixed $m < n$, define $\mathcal{P}_m$ to be the collection of matrices in $\R^{n\times m}$ with orthonormal columns. If $$P^*_m=\argmax_{P\in\mathcal{P}_m} \left(\min_{s\in S}\| P^Ts\|_{\ell_2}\right),$$ then $$\kappa_m =\min_{s\in S} \|(P^*_m)^Ts\|_{\ell_2}.$$
One may then construct a tuple of $\kappa_m$ values for a range of $m$. Such a tuple $(\kappa_{m_1},\kappa_{m_2},\ldots,\kappa_{m_\ell})$ is a $\kappa$-profile. Note that in this definition $m_1 \geq 1$ and $m_\ell \leq n$. 
\end{definition}
We remark that the set $\mathcal{P}_m$ is equivalent to the set of all orthogonal projections from $\mathbb{R}^n$ to $\mathbb{R}^m$.

Intuitively, if $\mathcal{M}$ is a manifold from which data is drawn, the $\kappa$-profile provides a measure of how successfully a projection $P^*$ embeds $\mathcal{M}$ into $\R^m$ for varying $m.$ Not only does this serve as a means of extracting information about the dimension of the underlying manifold, the $\kappa$-profile itself contains useful information about the nature of the data set. We demonstrate, for example, in Section~\ref{sect-weather} that the $\kappa$-profile reflects important features in data, such as the changing characteristics of weather data during the presence or absence of extreme weather events. 

Throughout this paper, we use the SAP algorithm to estimate the $\kappa$-profile in various settings; for simplicity we refer to the estimated profile as the $\kappa$-profile.

\subsection{Geodesic Distance on the Grassmann Manifold}
In this paper, we use the notion of geodesic distance on a Grassmann manifold to compare the projection returned by the SAP algorithm to the common dimensionality-reduction technique of principal component analysis. We briefly review the context of the Grassmannian and distance metrics on this manifold.

As described in \cite{kvinge2018too}, the {\sl Grassmannian} $Gr(k,n)$ is a manifold whose points parametrize the $k$-dimensional subspaces of a fixed $n$-dimensional vector space. An important feature of Grassmannians is that they can be given the structure of a differentiable manifold. One would like to determine the proximity of various points on a Grassmannian and this is typically carried out by first determining principal angles between the corresponding vector spaces. Principal angles between vector spaces are readily computable through a singular value computation as described below. 

Consider the subspaces $U$ and $V$ of a vector space $\mathbb R^n$ and let $q= \min\left\{\dim U,\dim V \right\}$.
The principal angles between $U$ and $V$ are the angles $\theta_1, \theta_2, \dots \theta_q \in [0,\frac{\pi}{2}]$ between pairs of principal vectors $\{ u_k, v_k \}$
with $u_1, \dots, u_q$ a distinguished orthonormal set of vectors in $U$ and $v_1, \dots, v_q$ a distinguished set of orthonormal vectors in $V$. These vectors are obtained recursively, for each $1\leq k \leq q$, by defining
\[
\cos \theta_k = \underset{u \in U, v \in V}{\max} u^T v = u_k^T v_k
\]
subject to 
\begin{itemize}
\item $||u||_2 = ||v||_2 = 1$
\item $u^T u_i = 0$ and $v^T v_i = 0$ for $i = 1,2, \ldots, k-1.$
\end{itemize}
A key point in this context is that any orthogonally invariant metric on a Grassmann manifold can be described as a function of principal angles. Furthermore, any function on pairs of points on a Grassmann manifold, that can be expressed as a function of the principle angles between the vector spaces corresponding to the points, will automatically be invariant under orthogonal transformations.

The principal angles and principal vectors between $U$ and $V$ can be determined from orthonormal bases for $U$ and $V$ as follows. Suppose $A$ (respectively $B$) are matrices whose columns form orthonormal bases for $U$ (respectively $V$). From the singular value decomposition we have a factorization $A^TB=Y\Sigma Z^T$. If $y_i$ (respectively $z_i$) denotes the $i^{th}$ column of $Y$ (respectively $Z$) then the $i^{th}$ singular vector pair can be computed as $u_i=Ay_i$ and $v_i=Bz_i$. Furthermore, the singular values of $A^TB$ are equal to $\cos \theta_1, \cos \theta_2, \dots, \cos \theta_q,$ where the sequence is assumed to be monotonically decreasing. See \cite{bjorck1973numerical}.

The geodesic distance between 2 points $X,Y$ on a Grassmann manifold $Gr(q,n)$ is defined in terms of the principal angles, $\theta_1, ..., \theta_q$, between the vector spaces represented by $X$ and $Y$ as $$d_{geodesic}(X,Y)=\sqrt{\theta_1^2+\theta_2^2+\cdots+\theta_q^2}.$$ While there are many other interesting orthogonally invariant metrics one could consider, for this paper we will utilize the geodesic distance.

\section{The $\kappa$-profile and time series}
\label{sect-dim-time-series}

The $\kappa$-profile can also provide information about the dynamics underlying data that is changing over time. Suppose that we are given a collection of data sets $\{D_t\}$ in $\mathbb{R}^n$ parametrized by time parameter $t \in \mathbb{Z}_{\geq 0}$, as well as a collection of bijective maps $f_t: D_t \rightarrow D_{t+1}$. Then we can identify points in $D_t$ and $D_{t+1}$, so that $f_t(x) \in D_{t+1}$ is the same point as $x \in D_{t}$, but one time step later. 

Given such a collection $\{D_t\}$ there are various ways of calculating the $\kappa$-profile depending on how we structure the input data. One approach is to create a $\kappa$-profile for each $D_t$ independently. That is, the secant set we apply our dimensionality reduction algorithm to is precisely the secant set of points $D_t \subset \mathbb{R}^n$ for each $t \in \mathbb{Z}_{\geq 0}$. This captures geometric properties of $\{D_t\}$ as $t$ changes. When we also want a single $\kappa$-profile to capture temporal changes in the data, an alternative approach is to construct a new sequence of data sets $\{D_t^{(\ell)}\}$ such that:
\begin{equation*}
D_t^{(\ell)} := \{[x,f_t(x),f_{t+1}(x), \dots, f_{t+\ell}(x)] \in \mathbb{R}^{n\ell} \; | \; x \in D_t \}
\end{equation*}
where the brackets indicate vector concatenation. This type of construction is often known as a {\emph{time-delay embedding}}. 

Motivation for this approach is given by Takens theorem \cite{Tak81}. Given a discrete dynamical system where the state space is an $m$-dimensional compact, smooth manifold $M$, evolution of the system can be defined as a smooth map $f: M \rightarrow M$. Suppose that we only have access to a single smooth measurement function $\varphi: M \rightarrow \mathbb{R}$. Takens theorem says that under suitably general conditions, we can construct a diffeomorphic copy of the manifold $M$ in $\mathbb{R}^k$ via a length $k \geq 2m+1$ time-delay embedding of $\varphi$ with respect to $f$, i.e. we define the embedding by sending $x$ to $[\varphi(x),f(\varphi(x)), \dots, f^k(\varphi(x))]$.

Hence when considering the problem of monitoring a large collection of data that is changing over time, it is useful to use time-delay embeddings so that we can not only monitor the current geometry of the data, but also changes in the dynamics of the data. In Section \ref{sect-weather}, we provide an example in which we use this framework to calculate $\kappa$-profiles for two weather data sets over time.  

Of course in the real world, we rarely know the dimension of the manifold $M$, as we generally only have access to some of the variables defining the dynamics of the system. It is an important problem therefore to estimate the value $2m+1$, or the shortest time-delay embedding that will fully capture the dynamics. Algorithms which address this problem include false nearest neighbors \cite{KBA92} and saturation methods \cite{GHKSS93}. In the spirit of the latter, the $\kappa$-profile should provide a new tool to approach this problem. Specifically, one should be able to estimate the value $2m +1$ by calculating the $\kappa$-profile for a data set $D_t^{(i)}$ for increasingly large $i$. When the embedding dimension is reached, change in the $\kappa$-profile for increasing $i$ should cease. We intend to explore this idea in future research.

\section{Example: the Kuramoto-Sivashinsky equation}
\label{sect-KS-equation}

The Kuramoto-Sivashinsky (KS) equation \cite{hyman1986kuramoto} in one spatial dimension is the partial differential equation 
\begin{equation*}
u_t + v u_{xxxx} + u_{xx} + \frac{1}{2}(u_x)^2 = 0,
\end{equation*}
where $v$ is a positive constant, $u: \mathbb{R} \times \mathbb{R}_{\geq 0} \rightarrow \mathbb{R}$ is a function that satisfies an $L$-periodic initial condition $u(x+L,0) = u(x,0)$, and each subscript on $u$ denotes a partial derivative with respect to that subscript. The KS equation was originally derived by Kuramoto for modeling the Belouz-Zabotinskii reaction \cite{KT75}. It was also developed by Sivashinsky to model thermal diffusive instabilities in laminar flame fronts \cite{MS77}. For $u(x,0)$ satisfying the initial condition, it has been proven that a unique solution exists and it is bounded. These solutions have the interesting characteristic that while they have coherent spatial structure, they exhibit temporal chaos. Furthermore, for different initial conditions and periodicity values $L$, the solution manifold has varying dimension. This makes data collected from the KS equation ideal for studying algorithms whose output should capture the dimension of a data set or related statistics.

Following \cite{hyman1986kuramoto}, the version of the KS equation that we use to generate data is
\begin{equation*}
u_t + 4u_{xxxx} + \alpha \Big(u_{xx} + \frac{1}{2}(u_x)^2 \Big) = 0,
\end{equation*}
where the periodicity has been set at $2\pi$, $v = 4$, and a bifurcation parameter $\alpha$ has been introduced. The stable solution manifolds for various values of $\alpha$ were described via numerical experiment in \cite{hyman1986kuramoto}. 

We now consider some examples with varying values of $\alpha$ and hence dimension. In each case, we generate a data set from the solution manifold consisting of 10,000 points in $\R^{32}.$ In several of these examples, we also provide a comparison with the PCA approach to dimensionality reduction for context.

\begin{figure}[h]
\includegraphics[width=8cm]{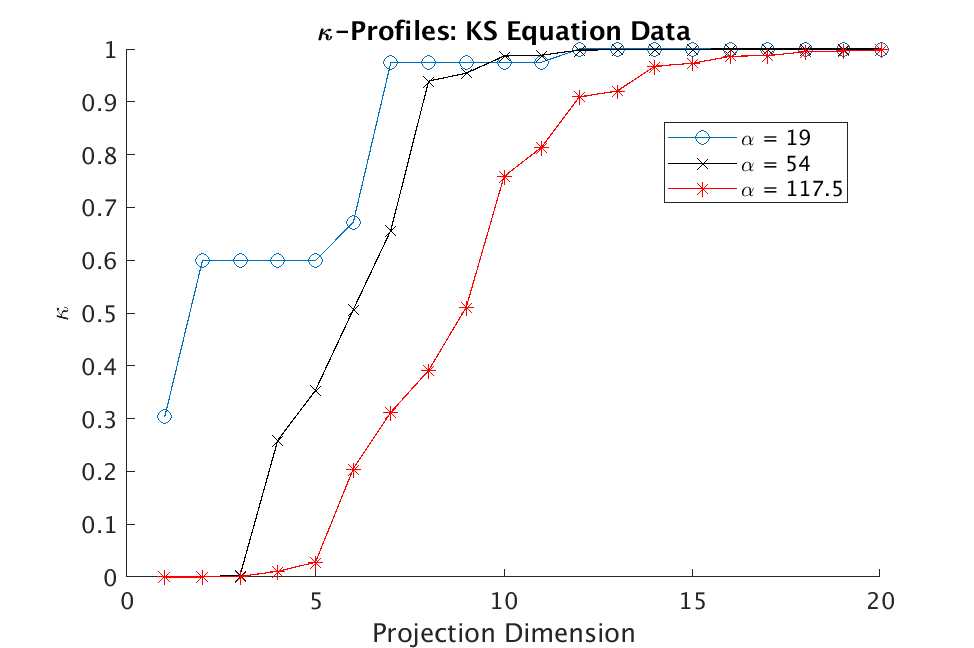}
\caption{\label{kappas} $\kappa$-profiles for the KS equation data with $\alpha=19, \alpha=54,$ and $\alpha=117.5$ for projection dimensions 1 to 20. Heuristically, if $\kappa\geq 0.2,$ we consider the embedding to be good.}
\end{figure}

Let us begin with $\alpha = 19.$ In this case, the authors of \cite{hyman1986kuramoto} note that the solution is a periodic orbit. In Figure~\ref{kappas}, we show the $\kappa$-profile for projection dimensions 1 to 20 for this value of $\alpha$ as well as several others. By heuristic, if the value of $\kappa$ is above 0.2, we consider the embedding to be good. Thus in this case, we can reasonably embed the data in $\R^1.$ It is worth noting as well that the projection into $\R^7$ is nearly isometric on the normalized secant set since the value of $\kappa$ is very close to one.

\begin{figure}[h]
\includegraphics[width=8cm]{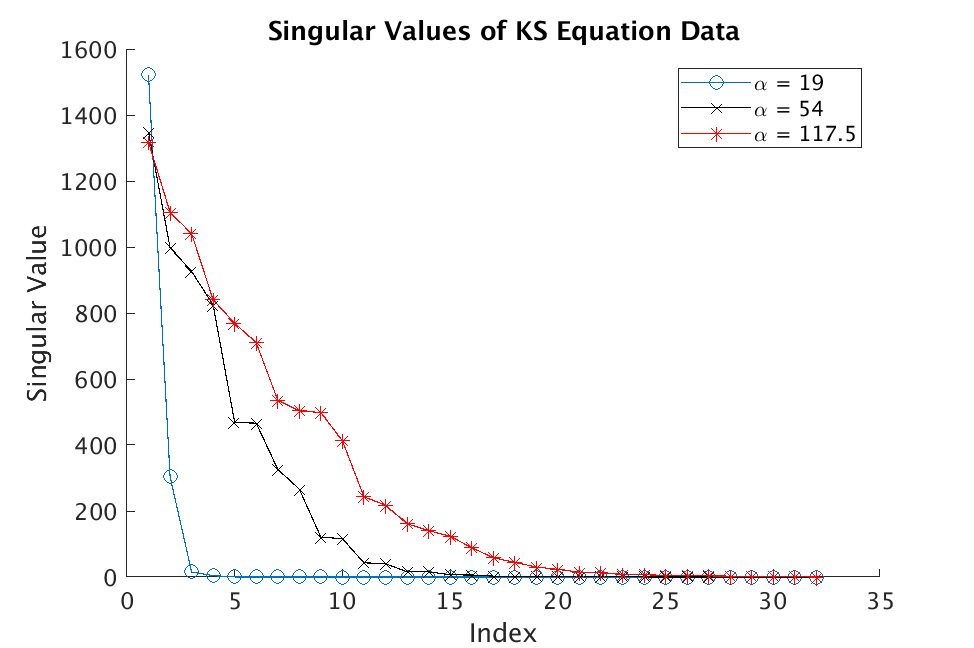}
\caption{\label{singulars} Singular values for the KS equation with $\alpha=19,\alpha=54,$ and $\alpha=117.5.$ In the case of $\alpha=19,$ the majority of the energy in the data is captured in the first one to two dimensions. For the other two choices of $\alpha,$ we see singular value decay but a less clear signal regarding the precise dimension of the data. Thus the added information from the $\kappa$-profile can be useful here.}
\end{figure}

\begin{figure}[h]
\includegraphics[width=8cm]{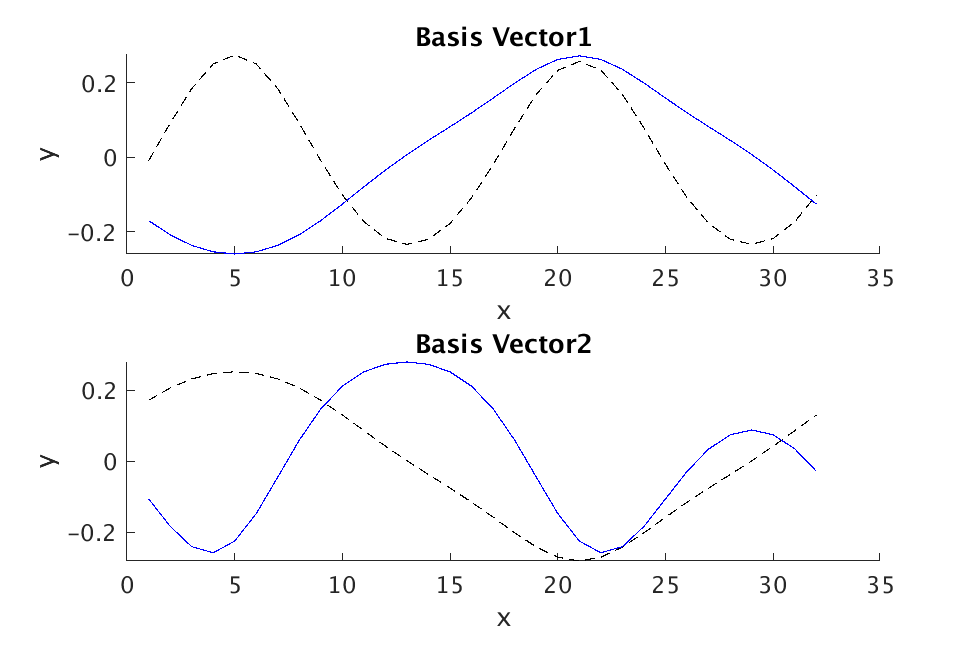}
\caption{\label{a19bases} For $\alpha=19:$ first two basis vectors for the two dimensionality-reduction techniques: first and second left singular vector for PCA (dashed, black line) and first and second basis vector from SAP (solid, blue line). Note that the SAP basis vectors are ordered in terms of $\kappa$ values.}
\end{figure}

\begin{figure}[h]
\includegraphics[width=8cm]{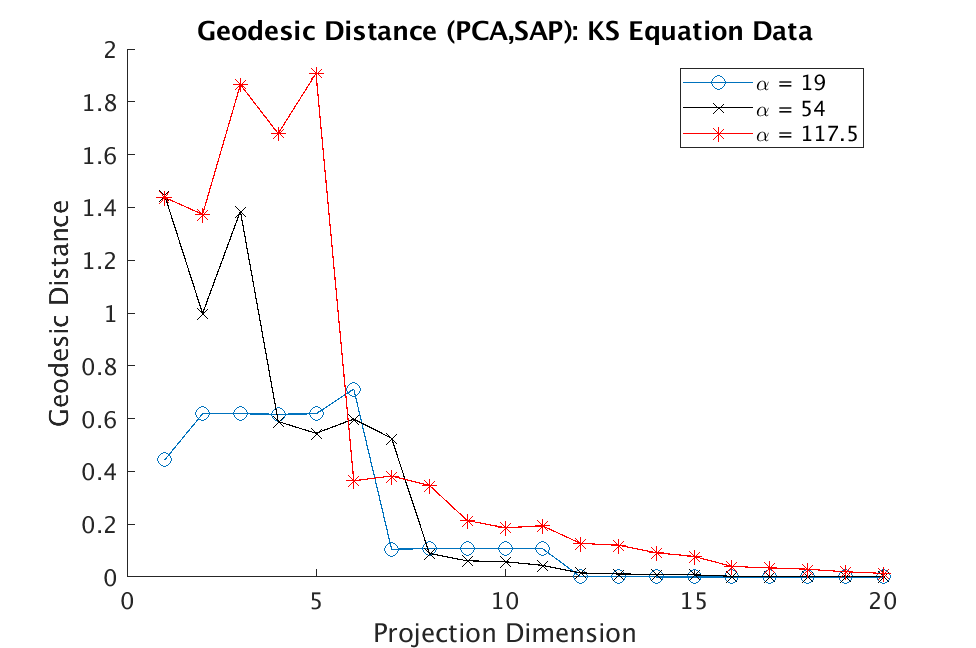}
\caption{\label{geodesics} Geodesic distance between basis of first $m$ left singular vectors and first $m$ SAP basis vectors for projection into $\R^m$ for KS equation with $\alpha=19,\alpha=54,$ and $\alpha=117.5$ for $m=1,\ldots,20.$ Note that the subspaces are quite similar for projection into $\R^{10}$ to $\R^{20}.$ However, the projections are fairly distinct for low dimensional projections.}
\end{figure}

For comparison, consider the singular values of the data set, shown in Figure~\ref{singulars}. In this example and throughout the paper, the data is not mean-subtracted. From the singular values for $\alpha=19$, we infer similar information to that contained in the $\kappa$-profile: the data can be projected into $\R^1$ or $\R^2$ without much loss of information. The two vectors that form a basis for the 2-dimensional subspace into which the data is projected for PCA and SAP are shown in Figure~\ref{a19bases}; while they capture similar information, they provide distinct projections. Note that the geodesic distance between the subspaces is approximately 0.6. For each embedding dimension, we consider the geodesic distance between the subspace that defines the PCA projection and that of the SAP projection; see Figure~\ref{geodesics}. Note that the subspaces are somewhat distinct for projections to $\R^m$ with $m<7.$

For a different choice of the parameter $\alpha,$ we see distinctly different behavior. For example, consider $\alpha = 54.$ The $\kappa$-profile for this $\alpha$ is shown in Figure~\ref{kappas}. In this example, we get a good projection of the data starting at dimension 4. For comparison, consider the singular values of the data shown in Figure~\ref{singulars}. Here, too, we see a change in behavior at dimension 4. However, the dimension estimate inferred from the singular values is not obvious - one could argue that the energy or variance in the data isn't essentially captured until some dimension between 10 and 15. Thus, the added information from the $\kappa$-profile is valuable. In this case, we see that we have a near isometry on the set of normalized secants when projecting into $\R^8.$ In \cite{hyman1986kuramoto}, this choice of $\alpha$ is stated to coincide with oscillatory and/or chaotic orbits.




As a final example from the KS equation data, we consider $\alpha=117.5,$ which produces chaotic orbits~\cite{hyman1986kuramoto}. We show the $\kappa$-profile in Figure~\ref{kappas} and the singular values in Figure~\ref{singulars}. Here we see that the more complex behavior of the solution set is reflected in the $\kappa$-profile: a good projection arises at dimension 6, and the $\kappa$ values increase more gradually than in the previous two examples with $\alpha=19$ and $\alpha=54.$ The jump between dimensions 6 and 7 in the singular values mirrors the information in the $\kappa$-profile. As before, the dimension estimate from the singular values is not obvious, so the $\kappa$-profile stands to provide meaningful additional context. 



We include in Figure~\ref{geodesics} a plot of the geodesic distance between the PCA and SAP projection subspace bases for dimensions 1 to 20 for all three choices of parameter $\alpha.$ Note that the PCA and SAP projections consistently differ in the low to intermediate dimensions.

\section{Example: weather data}
\label{sect-weather}

Statistics arising from measurements of the Earth's weather is one of the most fertile sources of big data available. In this section we provide an example of how large-scale regional weather patterns are reflected in the $\kappa$-profile of weather data sets. 

We obtained historical weather model data from the National Center for Environmental Prediction (NCEP) Climate Forecast System (CFS), version 2 \cite{CSFv2}. We took two rectangular grids, with data points at each $.5^{\circ}$ of latitude/longitude:
\begin{itemize}
\item The first is in the Western Atlantic ($20 - 33^{\circ}$N, $44 - 69^{\circ}$W), northeast of the Caribbean and in the path of Atlantic hurricane activity. The time frame for this data set is from April 1st 2014, to October 1st 2014 with measurements every 6 hours.
\item The second is located in the Western Pacific ($18-26^{\circ}$N, $123-132^{\circ}$E) east of Taiwan in the line of some of the Pacific typhoon activity. The time frame for this data set is from July 1st, 2015 to December 1st, 2015 with measurements every 6 hours.
\end{itemize}
The data contains $9$ weather variables at each grid point which include pressure, wind speed, air temperature, precipitation rate, etc. We consider points of these data sets to be the weather variables at a given grid location so that (before the time-delay embedding) the data sets consists of points in $\mathbb{R}^9$. We used a $19$ step time-delay embedding in order to construct new data sets $\{D_{A,t}\}$ (corresponding to the data from the Western Atlantic) and $\{D_{P,t}\}$ (corresponding to the data from the Western Pacific) from which we produced the $\kappa$-profiles that we analyze below. Note that for each time step $t$, corresponding to a 171 hour time window, $D_{A,t}$ and $D_{P,t}$ are sets of points in $\mathbb{R}^{171}$. As described in Section \ref{sect-dim-time-series}, one of the reasons for using a time-delay embedding is to capture the temporal dynamics of the entire weather system in this area.
 
\begin{figure*}
\includegraphics[width=18cm, height=6cm]{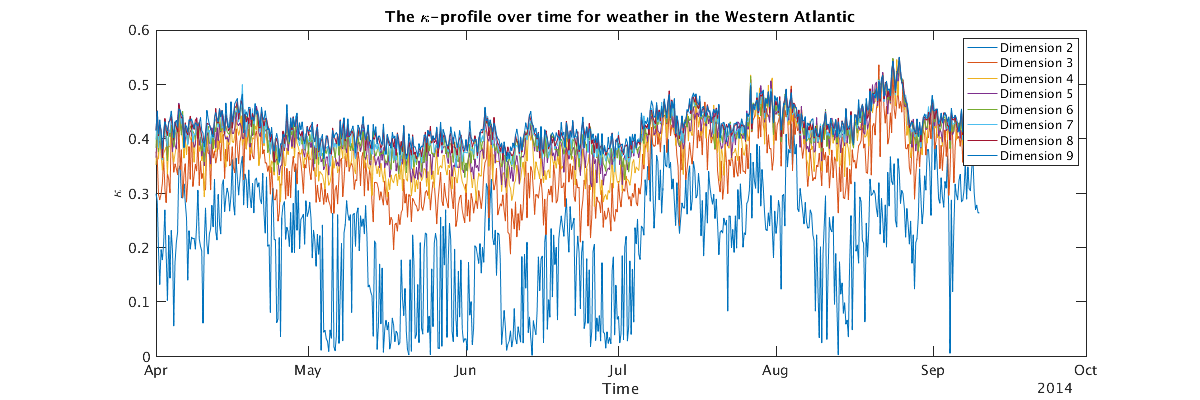}
\caption{\label{fig-west-atlantic} The $\kappa$-profile for weather data from the Western Atlantic in 2014 from the grid ($20 - 33^{\circ}$N, $44 - 69^{\circ}$W). The first hurricane of the year, Hurricane Arthur, approached the grid in the beginning of July, but because of the time-delayed embedding, the effect of the storm should be seen near the end of June. We suspect the drop in the $\kappa$-value corresponding to projection into 2 dimensions which occurs around the end of June is probably attributable to this. Similarly, the drop in the same $\kappa$-value in mid-August and early September are possibly related to Hurricane Cristobal and Hurricane Edouard respectively.}
\end{figure*}

\begin{figure*}
\includegraphics[width=18cm, height=6cm]{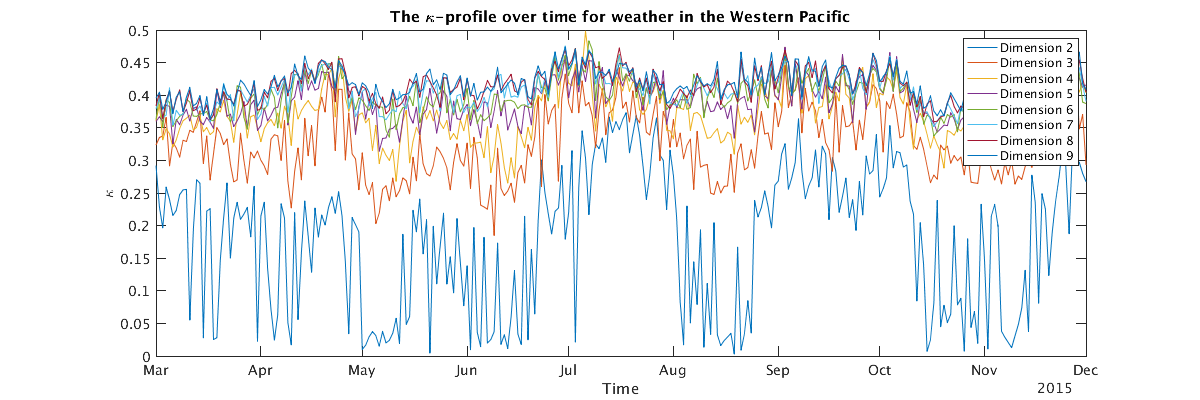}
\caption{\label{figure-west-pacific} The $\kappa$-profile for weather data taken from the rectangle ($18-26^{\circ}$N, $123-132^{\circ}$E) in the Western Pacific, east of Taiwan. We suspect that the sustained drop in the $\kappa$-value associated with projection into 2 dimensions beginning in May, may be related to Typhoon Dolphin which moved through the grid during this time (taking into account the time-delay embedding). The drop in $\kappa$-value at the beginning of August may possibly be attributed to Typhoon Soudelor.}
\end{figure*}

\begin{figure}[ht]
\includegraphics[width=8cm, height=6cm]{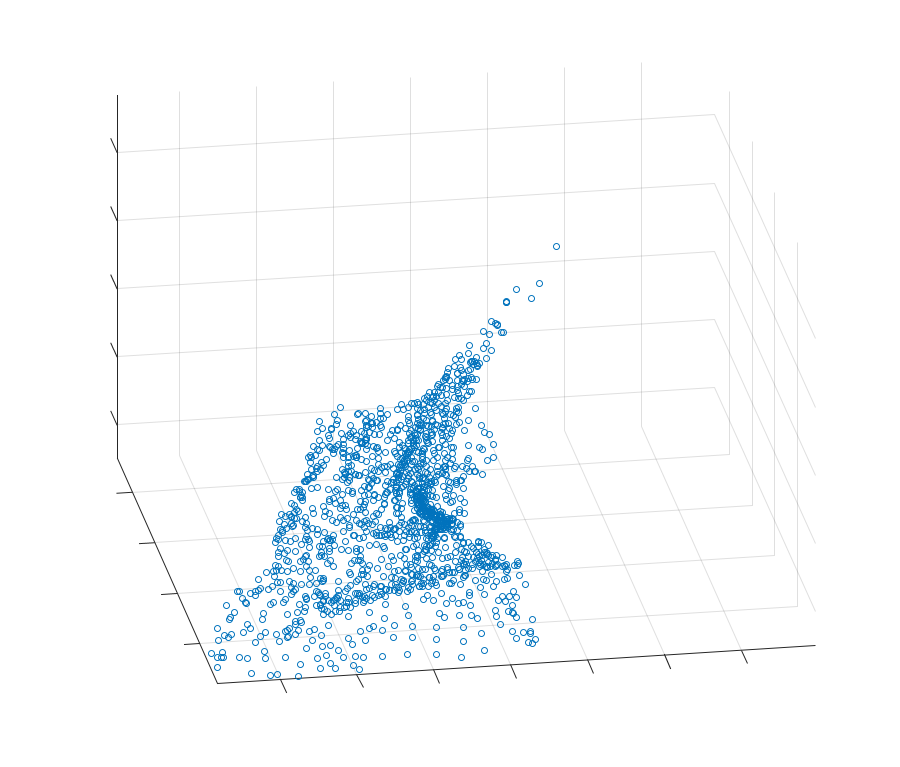}
\caption{\label{figure-proj3D-WA-363} Projection of points from an element of $\{D_{A,t}\}$ corresponding to a window beginning June 30th at 12:00pm, into $\mathbb{R}^3$. During this period, as indicated by Figure \ref{fig-west-atlantic}, the $\kappa$-value for projection into 2-dimensions is quite small, suggesting that the data is not 2-dimensional. This agrees with what can be seen from the 3-dimensional projection. We hypothesize that the disturbance of the dataset from 2-dimensions might be attributable to Hurricane Arthur that passed by during this time.} 
\end{figure}

\begin{figure}[ht]
\includegraphics[width=8cm, height=6cm]{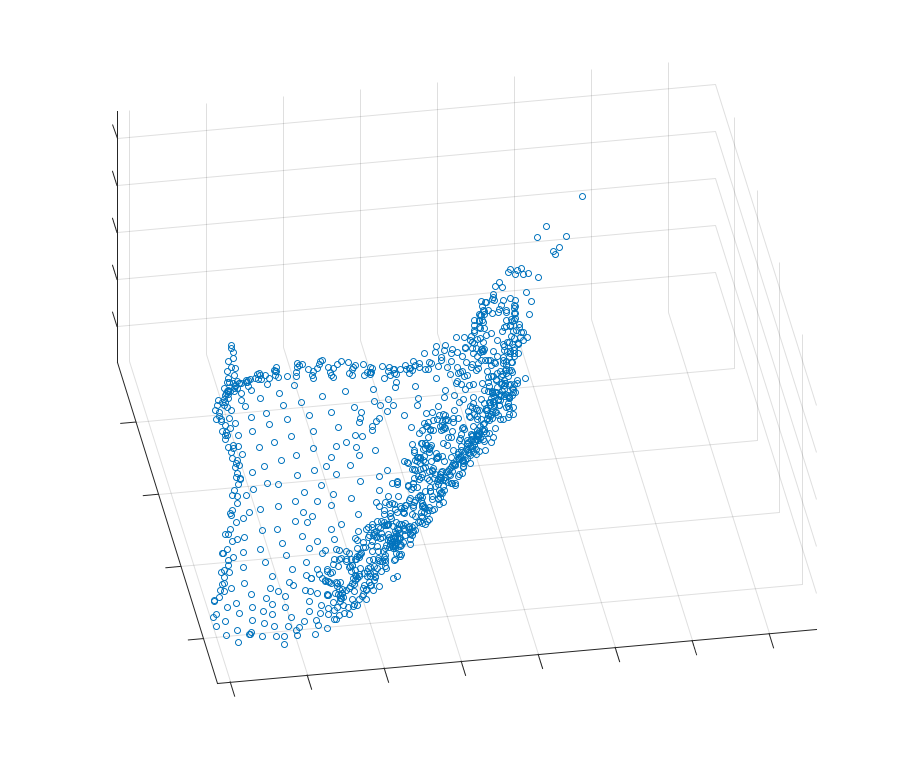}
\caption{\label{figure-proj3D-WA-410} Projection of points from an element of $\{D_{A,t}\}$ corresponding to a window of time beginning July 12th at 6:00am, into $\mathbb{R}^3$. Unlike Figure \ref{figure-proj3D-WA-363}, during this period the data set is very nearly 2-dimensional. This agrees with the $\kappa$-profile in Figure \ref{fig-west-atlantic}. There were no major storms in this window of time.}
\end{figure}

In Figures \ref{fig-west-atlantic} and \ref{figure-west-pacific} we provide $\kappa$-profiles for $\{D_{A,t}\}$ and $\{D_{P,t}\}$. We see that in both cases elements of $\{D_{A,t}\}$ and $\{D_{P,t}\}$ fluctuate between being 2 and 3 dimensional. This is further supported by examining projections of the data, obtained by the SAP algorithm (see Figures \ref{figure-proj3D-WA-363} and \ref{figure-proj3D-WA-410} for example). Rough comparisons of these figures with the record of storms (hurricanes and typhoons respectively) suggests that in general the $\kappa$-values decrease as a storm approaches (indicating an increase in the dimension of the data), and increase again when no major weather events are occurring. We hypothesize for example that the drop in $\kappa$-values in Figure \ref{fig-west-atlantic} in late-June/early-July is related to Hurricane Arthur which passed by the grid near this time (taking into account the time-delay embedding). It also seems likely that the drop in $\kappa$ values in mid-August is related to Hurricane Cristobal. 

In Figures \ref{figure-proj3D-WA-363} and \ref{figure-proj3D-WA-410} we project points from two elements of the collection $\{D_{A,t}\}$, from $\mathbb{R}^{171}$ into $\mathbb{R}^3$, using projections obtained from the SAP algorithm. Figure \ref{figure-proj3D-WA-363} corresponds to June 30th at 12:00pm, a time when Figure \ref{fig-west-atlantic} shows the $\kappa$-value for projection into 2-dimensions to be quite low. This is reflected in the shape of the data, which does not appear to be 2-dimensional. Figure \ref{figure-proj3D-WA-410} corresponds to a window beginning July 12th at 6:00am, a time where the $\kappa$-value for projection into 2-dimensions is quite high. Indeed, the points in this figure do appear to sit on a 2-dimensional plane.

While the $\kappa$-profiles shown in Figures \ref{fig-west-atlantic} and \ref{figure-west-pacific}  seem to capture historical weather activity in these areas during the given time range, the correlation does not seem to be perfect. For example, one would expect a drop in $\kappa$-values in Figure \ref{figure-west-pacific} during Typhoon Dujuan which passed through the grid in mid-September. It seems likely that such discrepancies could be understood if a direct connection between the physics of weather and changes in dimension of the corresponding data sets was known. We suggest this as an avenue for further research.

\begin{figure*}[!ht]
\includegraphics[width=18cm, height=6cm]{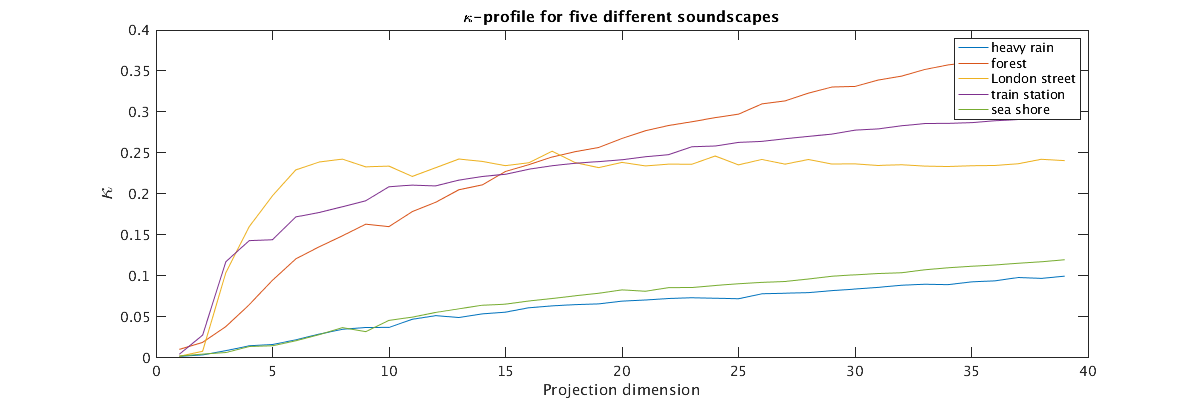}
\caption{\label{figure-soundscape-kappa} The $\kappa$-profile of $5$ soundscapes. The dimension of heavy rain and the sea shore appear to be much higher dimensional than the other environments.}
\end{figure*}

\section{Example: soundscapes}
\label{sect-soundscapes}

All environments produce ambient noise which serves as a continuous stream of information about the state of the particular location.

We ran the SAP algorithm on audio recordings from five different locations/states: (1) an area experiencing heavy rain (2) a forest, (3) a London street, (4) a train station, and (5) a sea shore \cite{FreeSound}. The audiofiles were originally recorded at a sampling rate of 48,000 with 2 channels. We resampled these at 1/100 this rate and took a point to be a length 5,000 window. For each environment then, the corresponding data set $D$ consists of some number of points in $\mathbb{R}^{10,000}$. We selected overlapping windows in order to capture a maximum amount of temporal information about the soundscape. 

We used the SAP algorithm to calculate the $\kappa$-profile for each of these soundscape data sets (Figure \ref{figure-soundscape-kappa}). We see that the heavy rain and sea shore soundscape appear to be quite high dimensional. This is perhaps unsurprising, as both of these soundscapes consist of relatively uniform, incoherent noise. On the other hand, the data sets for soundscapes corresponding to a train station and a London street appear to have lower dimension. This fits with the nature of urban ambient noise which generally has more structure than either the noise of a sea shore or heavy rain.

While we did not do it here, it would be easy to monitor a continuous sound recording in a manner similar to that found in Section \ref{sect-weather}. As indicated by Figure \ref{figure-soundscape-kappa}, such a set-up should capture fundamental changes in the soundscape through the $\kappa$-profile. For example, a sudden downpour in the London street environment would correspond to a sharp drop in $\kappa$-values in the $\kappa$-profile.

\section{Conclusion}

In this paper we present evidence suggesting that the $\kappa$-profile is a useful statistic for analyzing and monitoring data sets, particularly those that change over time. While it is in some sense a coarse statistic, it is broadly applicable and carries meaningful information.

We suggest a few directions for future research:
\begin{enumerate}
\item Given that dimension is an intrinsic property of a data set, it would be interesting to understand whether machine learning algorithms could benefit from inclusion of this as a feature.
\item It would be interesting to understand how the dimension and $\kappa$-profile of a data set is related to other notions of complexity of a data set such as entropy, etc.
\item As mentioned in Section \ref{sect-dim-time-series}, in the case where the data can be seen as reflecting an underlying dynamical system, the $\kappa$-profile should offer an alternative method of estimating the minimum embedding dimension. 
\end{enumerate}

\section*{Acknowledgment}

This paper is based on research partially supported by the National Science Foundation under Grants No. DMS-1513633, and DMS-1322508 as well as  DARPA awards  N66001-17-2-4020 and D17AP00004.

\bibliographystyle{IEEEtran.bst}
\bibliography{refs,KirbyFebruary2018,complete5}

\end{document}